\def\year{2022}\relax
\documentclass[letterpaper]{article} 
\usepackage{aaai22}  
\usepackage{times}  
\usepackage{helvet}  
\usepackage{courier}  
\usepackage[hyphens]{url}  
\usepackage{graphicx} 
\usepackage{natbib}  
\usepackage{caption} 

\DeclareCaptionStyle{ruled}{labelfont=normalfont,labelsep=colon,strut=off} 
\usepackage{multicol}
\usepackage{multirow}
\usepackage{booktabs}
\usepackage{tabularx}

\usepackage{algorithm}
\usepackage{algorithmic}
\usepackage[algo2e,ruled,vlined]{algorithm2e}
\usepackage{amssymb}
\usepackage{amsmath}
\usepackage{bm}
\usepackage{wrapfig}

\usepackage[implicit=false]{hyperref}
\usepackage{newfloat}
\usepackage{listings}
\floatstyle{ruled}
\newfloat{listing}{tb}{lst}{}
\floatname{listing}{Listing}
\title{Learning to Evolve on Dynamic Graphs}
\author {
    Xintao Xiang\textsuperscript{\rm 1}\equalcontrib,
    Tiancheng Huang\textsuperscript{\rm 2,3,4}\equalcontrib,
    Donglin Wang\textsuperscript{\rm 3,4} \thanks{Corresponding author.} 
}
\affiliations {
    \textsuperscript{\rm 1} Australian National University, Canberra, Australia 
    \textsuperscript{\rm 2} Zhejiang University, Hangzhou, China\\
    \textsuperscript{\rm 3} Westlake University, Hangzhou, China
    \textsuperscript{\rm 4} Westlake Institute for Advanced Study, Hangzhou, China\\
    xintao.xiang@anu.edu.au, \{huangtiancheng,wangdonglin\}@westlake.edu.cn
}

\usepackage{bibentry}

\begin{document}

\maketitle

\begin{abstract}
	Representation learning in dynamic graphs is a challenging problem because the topology of graph and node features vary at different time. 
	This requires the model to be able to effectively capture both graph topology information and temporal information.
	Most existing works are built on recurrent neural networks (RNNs), which are used to exact temporal information of dynamic graphs, and thus they inherit the same drawbacks of RNNs. 
	In this paper, we propose Learning to Evolve on Dynamic Graphs (LEDG) - a novel algorithm that jointly learns graph information and time information. 
	Specifically, our approach utilizes gradient-based meta-learning to learn updating strategies that have better generalization ability than RNN on snapshots. 
	It is model-agnostic and thus can train any message passing based graph neural network (GNN) on dynamic graphs. 
	To enhance the representation power, we disentangle the embeddings into time embeddings and graph intrinsic embeddings. 
	We conduct experiments on various datasets and down-stream tasks, and the experimental results validate the effectiveness of our method.
\end{abstract}

\section{Introduction}
\label{sec1}

Representation learning on graph data \cite{perozzi2014deepwalk,grover2016node2vec,kipf2016semi,velivckovic2017graph,GIN} has received increasing attention owing to its power in wide applications including finance, social networks and bioinformatics. However, most works focus on the static graph and ignore the fact that many real-world graphs are time-dependant. For example, in citation network, the graph is growing with time as more papers are published over time. In social network, the edges may even appear or disappear with time. In fact, learning representation among dynamic graphs is more challenging since 1) the features of graphs at different time may vary a lot even when the structures of graphs are similar so that the learned graph neural network (GNN) cannot generalize over time axis; 2) the topology of graph may change rapidly with time, which makes the model scalability becoming more crucial.

Recently, researches start to tackle the problems of representation learning on dynamic graphs \cite{goyal2018dyngem,skarding2020foundations}. Generally, the dynamic networks can be divided into two categories: \textit{discrete representations} which are a sequence of snapshots at different time intervals and \textit{continuous representations} which can be represented by graph streams, contact sequence or temporal events \cite{skarding2020foundations}. Our work falls into the category of representation learning on discrete representations of dynamic graphs. A main line of work in this category is based on recurrent neural networks (RNNs) such as GC-LSTM \cite{chen2018gc} and EvolveGCN \cite{pareja2020evolvegcn}. Typically, RNN-based methods learn to adjust the states (e.g. the weights of GCN \cite{kipf2016semi} in EvolveGCN \cite{pareja2020evolvegcn}) by the snapshots. Though the line of works achieves success, they have issues: 1) they suffer from same issues of RNNs that they cannot compress long-range dependencies into hidden states and they cannot be paralleled \cite{bahdanau2015neural}; 2) Methods like EvolveGCN \cite{pareja2020evolvegcn} is similar to model-based meta-learning methods which use RNN to update the model parameters but recent research have found that such methods are more likely to overfit and have limited generalization ability compared to gradient-based meta-learning methods \cite{finn2017meta}. 

Aiming to address the issues mentioned above, we propose to use gradient-based meta-learning \cite{finn2017model} for dynamic graphs. Besides, to enhance the representation power of the model, we propose to disentangle the embedding to capture time information and graph information. A key intuition of this approach is that time affects significantly on both the physical meaning of nodes (e.g. a researcher worked on statistics 10 years ago but now works on deep learning) and the task target such as link prediction (e.g. customers in year 2021 are more likely to buy electrical-powered cars than year 2011). As a result, the time biases our training objective significantly. We thus argue that: a) the embeddings of dynamic graphs are formed by \textit{time information} and \textit{graph intrinsic information} (graph structure and attributes of nodes), where b) time information continuously changes with time and gives a prior on prediction targets, and graph intrinsic information directly contributes to the prediction targets. 

Based on the discussions above, we propose a novel algorithm Learning to Evolve on Dynamic Graphs (LEDG). First, we formulize our argument above by explicitly disentangling the embedding into \textit{time embedding} and \textit{graph intrinsic embedding}. The final prediction is performed by the combination of predictions on time embedding and graph intrinsic embedding. As the relative time between snapshots can be observed, we use a time predictor to predict the time by time embeddings to make sure that the embddings capture the time information. Second, we borrow the idea of gradient-based meta-learning \cite{finn2017model} and use episodic training to learn a model with the best initialization parameters that can quickly adapt to future graphs with only a small number of historical graphs. Our algorithm is model-agnostic and can be used for any message passing based GNN even if it is designed for static graphs in nature. 
Our main contributions are as follows:

 (1) We propose a simple but effective attention-based method to disentangle the embeddings of dynamic graphs into time embeddings and graph intrinsic embeddings. 
 
 (2) We propose a novel algorithm LEDG based on gradient-based meta-learning and can train any message passing based GNN on dynamic graphs.
 
 (3) We perform detailed experiments of our algorithm on various datasets and the results indicate that our algorithm help base model get higher performance.


\section{Related Work}
In this section, we summarily introduce some related work about static graph representation learning, and dynamic graph representation learning.

\subsection{Static Graph Representation Learning} 
Over the years, various deep learning based methods have been proposed to learn representations on static graphs. Early attempts of embedding learning on graphs are inspired by Skip-gram \cite{mikolov2013distributed} and are based on random walks such as DeepWalk \cite{perozzi2014deepwalk} and Node2Vec \cite{grover2016node2vec}. Another class of graph representation learning is based on message passing (i.e. neighborhood aggregation) such as methods proposed in \cite{kipf2016semi,velivckovic2017graph,hamilton2017inductive}. Though the two classes of methods have achieved great success in representing graph-structured data, most of them are designed for static graphs in nature and cannot be directly applied on evolving graphs with temporal information. In this paper, with our proposed framework, static methods based on message passing can be applied in the dynamic setting to capture both graph topology and temporal information.

\subsection{Dynamic Graph Representation Learning} 
Dynamic graph representation learning aims to deal with a more challenging problem that the graph is changing over time. \cite{skarding2020foundations,kazemi2020representation} are two surveys about researches on dynamic graphs. Discrete representation of dynamic graphs is widely used to represent a dynamic network by a sequence of snapshots where each snapshot represents the network in a specific time interval. Various algorithms have been proposed to tackle representation learning on such discrete dynamic graphs \cite{li2019predicting,sankar2020dysat}. Combining RNN and GNN is an intuitive idea and a branch of researches have made attempts based on this idea \cite{chen2018gc,jin2019recurrent,manessi2020dynamic}. For example, EvolveGCN \cite{pareja2020evolvegcn} integrates RNN into graph convolutional network (GCN), where for each snapshot, the weight of GCN is encoded by RNN according to the historical information. In this paper, we focus on discrete dynamic networks. Different from the stated previous works which design models specific for dynamic graphs, we propose a generic algorithm that can adapt GNNs on dynamic setting even if they are typically designed for static graphs.

\section{Preliminaries}
\label{section2}

\subsection{Notations} 

$\mathcal{G}^t=(\mathcal{V}^t, \mathcal{E}^t)$ denotes a graph with nodes $\mathcal{V}^t$ of features $\mathbf{X}^t$, and undirected edges $\mathcal{E}^t$ with adjacency matrix $\mathbf{A}^t$ at time $t$. The dynamic graph can be represented by a time-ordered sequence of graphs $\mathbb{G}=\{\mathcal{G}^{1}, \mathcal{G}^{2},...,\mathcal{G}^{T}\}$, where $\mathcal{G}^{1}$ is the initial snapshot and $\mathcal{G}^T$ is the last snapshot at time $T$. With the graph evolving, the number of nodes and the number of edges may increase or decrease. Let $\mathbf{H}^t$ represent the embeddings at time $t$, where the initial embeddings come from $\mathbf{X}^t$ (to simplify the notations, otherwise denoted, $\mathbf{X}$ and $\mathbf{H}$ represent embeddings at time $t$).

\subsection{Recall Static GNNs} 
Recent years have witnessed the success of GNNs for relational data \cite{kipf2016semi,velivckovic2017graph}. Our proposed methods lay on utilizing GNNs which are built on message passing, where the local neighborhood information is aggregated iteratively to get more contextual representation, to be fit in a dynamic setting. Generally, a message passing based graph neural network (MPGNN) can be represented as 
\begin{equation}
  \mathbf{h}^{(l)}_v = \sigma\Big(\mathop{Aggregate^{(l)}}\limits_{\forall u\in\mathcal{N}(v)}\big(f(\mathbf{h}^{(l-1)}_u, \mathbf{A}_{vu})\big) \Big),
\end{equation}
where $\mathbf{h}^{(l)}_v$ represents the representation of node $v$ at layer $l$, $\sigma$ is the (non-linear) activation function and $\mathcal{N}(v)$ denotes the neighborhood of node $v$ (with or without self-loop). $Aggregate$ including $sum$, $max$ or $mean$ gathers the information from neighborhood. $f(\cdot)$ denotes a function that can extract the information of nodes. 

Various GNNs can be represented in this form. An example of this neural network is GCN \cite{kipf2016semi}, in which the message passing function can be represented as 
\begin{equation}
\mathbf{h}^{(l)}_v = ReLU\big(\mathop{Sum}\limits_{\forall u \in \mathcal{N}(v)}(\widehat{\mathbf{A}}_{vu}\mathbf{h}^{(l-1)}_u\mathbf{W}^{(l)})\big)
\end{equation}
where 
\begin{equation}
\widehat{\mathbf{A}}=\widetilde{\mathbf{D}}^{-\frac{1}{2}}\widetilde{\mathbf{A}}\widetilde{\mathbf{D}}^{-\frac{1}{2}},\ \widetilde{\mathbf{A}}=\mathbf{A}+\mathbf{I},\ \widetilde{\mathbf{D}}=diag\big(\mathop{\sum}\limits_{j}\widetilde{\mathbf{A}}_{ij}\big).
\end{equation}
These methods have a basic assumption (homophily \cite{Geom-GCN}) that nearby nodes are similar so that aggregating the information from neighbors can enrich the information of nodes. By utilizing this framework, nodes aggregate useful information from neighbors and the final representations can be directly used for the down-stream tasks.

\subsection{Problem Definition} 
Given a dynamic graph $\mathbb{G}=\{\mathcal{G}^{1}, \mathcal{G}^{2},...,\mathcal{G}^{T}\}$, representation learning on this graph with $T$ snapshots aims to learn representations $\mathbf{H}^t$ of nodes at time $t=t_0,t_1,...,T$, such that $\mathbf{H}^t$ can preserve both time information and graph intrinsic information, and thus can be used for down-stream tasks such as link prediction, edge classification, and node classification in future time.

\section{Proposed Method}
\begin{figure}
    \centering
	\includegraphics[width=0.47\textwidth]{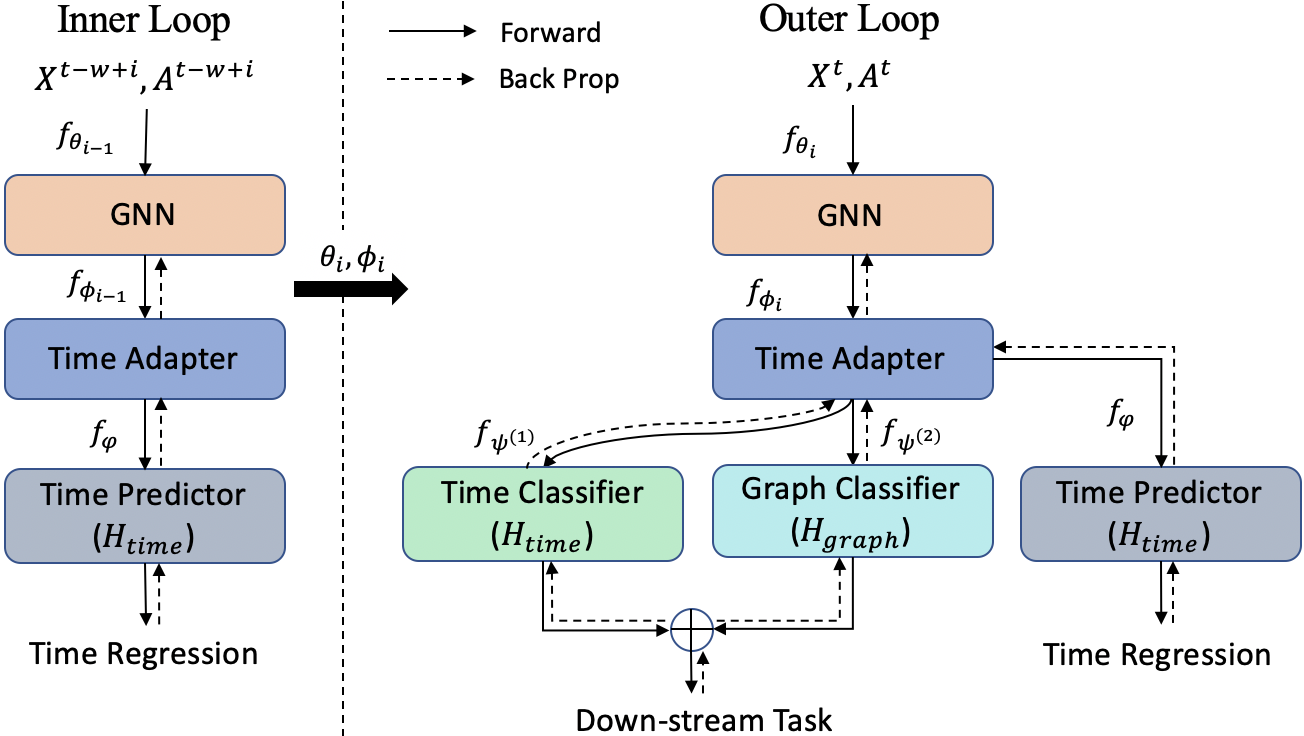}\vspace{-3pt}
	\caption{An illustration of our method. In inner loop, only GNN and time adapter are updated, while in outer loop, all the parameters are optimized.}\vspace{-8pt}
	\label{framework}
\end{figure}

In this section, we formally introduce how we disentangle the graph embeddings into time embeddings and graph intrinsic embeddings respectively, and how we train the model in a gradient-based meta-learning way. An overview of the method can be seen in Fig. \ref{framework}.

\subsection{Feature Disentanglement}

In this subsection, we introduce how we disentangle the embeddings and how we calculate the losses in one snapshot. We regard each snapshot as a static graph. Then the initial embeddings of the current snapshot are encoded by a message passing based GNN denoted by $f_{\theta}$. Mathematically, for a snapshot with node features $\mathbf{X}$ and adjacency matrix $\mathbf{A}$, the embeddings are encoded as:
\begin{equation}
\mathbf{H}=f_{\theta}(\mathbf{X},\mathbf{A}),
\end{equation}
where $\mathbf{H}\in \mathbb{R}^{N \times D}$, $N$ is the number of nodes and $D$ denotes the hidden dimension. As mentioned in Section \ref{sec1}, $\mathbf{H}$ is a mixture of graph information and time information, which we aim to disentangle.

\noindent\textbf{Feature disentanglement on dynamic graph.}
We assume $\mathbf{H}= \mathbf{H}_{graph} + \mathbf{H}_{time}$, where $\mathbf{H}_{graph} \in \mathbb{R}^{N \times D}$ denotes the graph intrinsic embeddings and $\mathbf{H}_{time}\in \mathbb{R}^{N \times D}$ denotes the time embeddings. This assumption is reasonable as the original embeddings come from both the time and the graph. Given embeddings $\mathbf{H}$, we employ feature-wise attention to disentangle them into $\mathbf{H}_{graph}$ and $\mathbf{H}_{time}$. A time adapter $f_{\phi}$ which is a multilayer perceptron (MLP) is used to get the attention map $\mathbf{S} \in \mathbb{R}^{N \times D}$ by:
\begin{equation}
\mathbf{S} = \sigma\big(f_{\phi}(\mathbf{H})\big),
\end{equation}
where $\sigma$ represents $Sigmoid$ function. The time embeddings $\mathbf{H}_{time}$ and graph intrinsic embeddings $\mathbf{H}_{graph}$ are then calculated by:
\begin{align}
\mathbf{H}_{graph} &= \mathbf{S} \odot \mathbf{H},\\  \mathbf{H}_{time} &= (\mathbf{1} - \mathbf{S}) \odot \mathbf{H},
\end{align}
where $\odot$ denotes Hadamard Product. By taking this attention based operation, we divide the original embedding $H$ to time embedding and graph intrinsic embeddings at every dimension of embedding.

\noindent\textbf{Time regression.} Recall that we expect the model to be able to recognize what time position the current snapshot is in. To restrict $\mathbf{H}_{time}$ to best represent the temporal information, we use a time predictor denoted by $f_{\varphi}$ which is an MLP to predict the current time $t$ by $\mathbf{H}_{time}^t$ in this snapshot. We formulate the loss as:
\begin{equation} \small
	\label{loss_time}
	\begin{split}
	\mathcal{L}_{time}\big(\mathbf{H}_{time}^t; f_{\theta}, f_{\phi}, f_{\varphi}\big) = smooth_{L_1}\big(f_{\varphi}(Pool(\mathbf{H}_{time}^t))- t\big),\\
in\ which \quad
smooth_{L_1}(x) = \left\{
	\begin{array}{lcl}
	0.5x^2       &      & {if \  |x|<1}\\
	|x| - 0.5    &      & {otherwise}
	\end{array} \right.
\end{split}
\end{equation}
is a robust $L_1$ loss that is less sensitive to outliers than $L_2$ loss. $Pool$ denotes graph pooling. In this paper for simplicity, we use mean-pooling and so the mean of $\mathbf{H}_{time}$ is used in the calculation.

\noindent\textbf{Down-stream tasks.} As the final prediction is related to both the time and the graph, we use two classifiers $f_{\psi^{(1)}}$ and $f_{\psi^{(2)}}$ which are two MLPs to calculate the predictions for down-stream tasks by:
\begin{equation}
\tilde{\mathbf{Y}} = Softmax(f_{\psi^{(1)}}(\mathbf{H}_{time}) + f_{\psi^{(2)}}(\mathbf{H}_{graph})).
\end{equation}
Note that following the setting in \cite{pareja2020evolvegcn}, for link prediction and edge classification, $\mathbf{H}_{time}$ and $\mathbf{H}_{graph}$ will be reformulated by the concatenation of source and target node embeddings. Then the final loss can be simply calculated by cross-entropy loss in the form by:
\begin{equation} \small
\label{loss_task}
	\mathcal{L}_{task}\big(\mathbf{H}_{time},\mathbf{H}_{graph};f_{\theta}, f_{\phi}, f_{\psi}\big) = CrossEntropy \big(\tilde{\mathbf{Y}}, \mathbf{Y}\big),
\end{equation}
where $\mathbf{Y}$ represents the ground truths of tasks (i.e. edge existence in link prediction task, and node labels in node classification task).

\begin{algorithm2e}[t]
	\caption{Learning to Adapt to Evolving Graphs} 
	\label{alg:LAEG}
	\SetAlgoLined
	\LinesNumbered  
	\KwIn{Graph: $\mathbb{G}=\{ \mathcal{G}^1,...,\mathcal{G}^T\}$; Inner loop learning rate: $\eta_{in}$; Outer loop learning rate: $\eta_{out}$; GNN model: $f_{\theta}$; Time Adapter: $f_{\phi}$; Time predictor: $f_{\varphi}$; Classifier: $f_{\psi}$}
	\While {not done} {
		\For {$t$ $in$ $[w : T]$}
		{
			Initialize $(\theta_0,\phi_0) \leftarrow (\theta, \phi)$\\
			\For {$i$ $in$ $[1 : w]$}{
				In inner loop, evolve the parameters of GNN $f_{\theta}$ and adapter $f_{\phi}$ by Equation \ref{innerloop} where \\
				$(\theta_{i},\phi_{i})= (\theta_{i-1},\phi_{i-1}) - \eta_{in}\big[\nabla_{(\theta,\phi)}\big(\mathcal{L}_{time}\big(\mathbf{H}_{time}^{t-w+i}; f_{\theta_i}, f_{\phi_i}, f_{\varphi}\big)\big)\big]$
			}
			In outer loop, update all the parameters by Equation \ref{outerloop} where \\
			$(\theta,\ \phi, \psi, \varphi) \leftarrow (\theta, \phi, \psi, \varphi)\ - \eta_{out}\nabla_{(\theta, \phi, \psi, \varphi)}$
			$\Big[\sum_{i=1}^{w} \Big( \mathcal{L}_{task}\big(\mathbf{H}_{time}^t,\mathbf{H}_{graph}^t;f_{\theta_i}, f_{\phi_i}, f_{\psi}\big) + 
			\lambda \mathcal{L}_{time}\big(\mathbf{H}_{time}^t; f_{\theta_i}, f_{\phi_i}, f_{\varphi}\big) \Big) \Big]$
		}
	}
\end{algorithm2e}

\subsection{Meta Framework}

In this subsection, we describe how we adopt the meta-learning training strategy \cite{finn2017meta} on the dynamic graphs. \cite{finn2017meta} formulates tasks in few-shot learning setting, where the training objective is to learn a best initialization of model with a best updating rules that can quickly adapt to few samples with good generalization ability while in dynamic graph setting, the objective is to learn updating strategies that can quickly adapt to a sequence of historical snapshots. The time range of the sequence is denoted by a time window of size $w$. An overview of the algorithm can be seen in Algorithm \ref{alg:LAEG}.

\noindent\textbf{Adapt feature extractor in inner loop.} The objective of inner loop is to evolve the parameters of feature extractor (i.e. GNN and time adapter) according to a sequence of historical snapshots (see Fig. \ref{framework} left). We adapt the feature extractor by time regression. Specifically, for predicting snapshot at time $t$, we first initialize parameters $(\theta_0,\phi_0) \leftarrow (\theta,\phi)$ by the current model. Then we use SGD to update the parameters through $w$ closest snapshots by Equation \ref{loss_time}. Denote index of snapshot in a time window as $i=1,...,w$, the exact time of the snapshot is then $t-w+i$. Formally,
\begin{equation}
	\label{innerloop}
	\begin{aligned}
		(\theta_{i},\phi_{i}) &\leftarrow\ (\theta_{i-1},\phi_{i-1})\\ & - \eta_{in}\big[\nabla_{(\theta,\phi)}\big(\mathcal{L}_{time}\big(\mathbf{H}_{time}^{t-w+i}; f_{\theta_{i-1}}, f_{\phi_{i-1}}, f_{\varphi}\big)\big)\big], 
	\end{aligned}
\end{equation}
where $\eta_{in}$ is the inner loop learning rate. Note that as we re-index graphs in the time window thus in calculating $\mathcal{L}_{time}$, the prediction is from $i$ to $w$ rather than $t-w+i$ to $t$.

\noindent\textbf{Update all parameters in outer loop.} The objective of outer loop is to make the adaptation in inner loop more effective where each update should bring better performances on target snapshot $\mathcal{G}^t$. In each update step $i$ of inner loop, we evaluate its performance on our target snapshot by Equation \ref{loss_time} and Equation \ref{loss_task}, and formally,
\begin{equation} \small
	\label{outerloop}
	\begin{aligned}
	  (&\theta,\ \phi, \psi, \varphi) \leftarrow (\theta, \phi, \psi, \varphi)\ \\ 
	  & - \eta_{out}\nabla_{(\theta, \phi, \psi, \varphi)}\Big[\sum_{i=1}^{w} \Big( \mathcal{L}_{task}\big(\mathbf{H}_{time}^t,\mathbf{H}_{graph}^t;f_{\theta_i}, f_{\phi_i}, f_{\psi}\big) \\
	  &+ \lambda \mathcal{L}_{time}\big(\mathbf{H}_{time}^t; f_{\theta_i}, f_{\phi_i}, f_{\varphi}\big) \Big) \Big],
	\end{aligned}
\end{equation}
where $\eta_{out}$ denotes the outer loop learning rate and $\lambda$ is a hyperparameter that is used to balance the two losses. The optimized model parameters will be the initial parameters of next inner loop.

\section{Experiment}




In the following sections, we provide the dataset description, compared methods, and evaluation metric.
\subsection{Datasets}

We verify our method on seven publicly available datasets. Each dataset contains a sequence of time-ordered graphs. We follow the dataset preprocessing and splitting setting of the datasets that are used in \cite{pareja2020evolvegcn}. The brief descriptions of the datasets are as follows:

\begin{itemize}
  \item \textbf{Stochastic Block Model\footnote{\href{https://github.com/IBM/EvolveGCN/tree/master/data}{https://github.com/IBM/EvolveGCN/tree/master/data}} (SBM)}: SBM is a random graph model for simulating community structures and evolutions. The SBM we used in the experiment is the one which is generated by \cite{pareja2020evolvegcn}.
  
  \item \textbf{Bitcoin OTC\footnote{\href{http://snap.stanford.edu/data/soc-sign-bitcoin-otc.html}{http://snap.stanford.edu/data/soc-sign-bitcoin-otc.html}} (BC-OTC)}: This is a network of a platform where people trade Bitcoin. The edges are the rates that members give other members in a scale of -10 (total distrust) to +10 (total trust).

  \item \textbf{Bitcoin Alpha\footnote{\href{http://snap.stanford.edu/data/soc-sign-bitcoin-alpha.html}{http://snap.stanford.edu/data/soc-sign-bitcoin-alpha.html}} (BC-Alpha)}: The network is similar to BC-OTC but people in this network trade Bitcoin on a different platform.

  \item \textbf{UC Irvine Messages\footnote{\href{http://networkrepository.com/opsahl\_ucsocial.php}{http://networkrepository.com/opsahl\_ucsocial.php}} (UCI)} \cite{nr}: This network is a social network where the nodes represent online community of students in the University of California and edges represent sent messages.

  \item \textbf{Autonomous Systems\footnote{\href{http://snap.stanford.edu/data/as-733.html}{http://snap.stanford.edu/data/as-733.html}} (AS)}: This is a communication network where each router exchanges traffic flows with some neighbors. 

  \item \textbf{Reddit Hyperlink Network\footnote{\href{http://snap.stanford.edu/data/soc-RedditHyperlinks.html}{http://snap.stanford.edu/data/soc-RedditHyperlinks.html}} (Reddit)}: The network represents the links from one post in the source community to another post in the target community. The dataset contains computed embeddings.

  \item \textbf{Brain}\footnote{\href{https://tinyurl.com/y6d74mmv}{https://tinyurl.com/y6d74mmv}} \cite{xu2019adaptive}: The nodes in this network represent tiny cubes of brain issues and the edges indicate the cubes' connectivity. Different from \cite{xu2019adaptive} that uses all snapshots to train, our task is more challenging as we have no access to val/test snapshots during training.
\end{itemize}

For a fair comparison, we follow the dataset preprocessing and splitting setting of datasets that are used in \cite{pareja2020evolvegcn}. The summarized datasets are displayed in Table \ref{tab:dataset}.

\begin{table*}[t] \small
	\centering
	\caption{Statistics of datasets.}
    \begin{tabular}{lccccccc}
      \hline
      Datasets & SBM & BC-OTC & BC-Alpha & UCI & AS & Reddit & Brain \\
      \midrule
      \#Nodes & 1,000 & 5,881 & 3,777 & 1,899 & 6,474 & 55,863 & 5,000\\
      \#Edges & 4,870,863 & 35,588 & 24,173 & 59,835 & 13,895 & 858,490 & 1,955,488\\
      \#Time Splits (Train/Val/Test) & 35/5/10 & 95/14/28 & 95/13/28 & 62/9/17 & 70/10/20 & 122/18/34 & 10/1/1\\ 
      \#Tasks: Link Prediction & \checkmark & \checkmark & \checkmark & \checkmark & \checkmark &  $\times$ & $\times$ \\ 
      \#Tasks: Edge Classification & $\times$ & \checkmark & \checkmark & $\times$ & $\times$ & \checkmark & $\times$ \\
      \#Tasks: Node Classification & $\times$ & $\times$ & $\times$ & $\times$ & $\times$ & $\times$ & \checkmark \\
      \hline
	  \end{tabular}%
	\label{tab:dataset}%
\end{table*}%

\subsection{Compared Methods}

\noindent\textbf{Baselines.} To validate the effectiveness of our method, we compare two pairs of baselines: (1) \textit{Static graph representation learning methods}, including the most commonly used GCN \cite{kipf2016semi} and GAT \cite{velivckovic2017graph}, which are trained on every snapshot and accumulate the gradients over the time axis; and (2) \textit{Dynamic graph representation learning methods}, including GCN-GRU which is implemented the same as the one used in \cite{pareja2020evolvegcn}, EvolveGCN \cite{pareja2020evolvegcn} with its two variants EvolveGCN-O and EvolveGCN-H, DynGEM \cite{goyal2018dyngem}, and dyngraph2vec \cite{goyal2020dyngraph2vec} with its two variants dyngraph2vecAE (V1) and dyngraph2vecAERNN (V2).

\noindent\textbf{Variants of our proposed method.} Our method can fit to any message passing based GNN. We test the performance by using the two most popular GCN \cite{kipf2016semi} and GAT \cite{velivckovic2017graph}. For all experiments of our methods, we use $\lambda=0.1$. The hyperparameters are tuned by grid search where the hidden size is selected from [32, 64, 128, 256] and the outer loop learning rate $\eta_{out}$ is chosen from [0.001, 0.002, 0.005]. The inner loop learning rate $\eta_{in}$ is 10 times the corresponding outer loop learning rate. The time adapter, time predictor and classifier are all two-layer MLPs with ReLU as the activation function.

\subsection{Evaluation Metric} 
In this paper, we consider three down-stream tasks such as link prediction, edge classification, and node classification on dynamic graphs.
For link prediction, we use Mean Average Precision (MAP) and Mean Reciprocal Rank (MRR), which are two widely adopted metrics and can be used in dynamic link prediction tasks \cite{pareja2020evolvegcn,skarding2020foundations}. For edge classification and node classification, we both use \emph{Micro-$F_1$} as evaluation metric.

\begin{table}[t] \small
	\centering
	\caption{Link prediction results where mean average precision (MAP) and mean reciprocal rank (MRR) are displayed.}\vspace{-3pt}
	\scalebox{0.8}{
	\begin{tabular}{c|cc|cc|cc}
	\hline
	\multirow{2}{*}{}Datasets&
	\multicolumn{2}{c|}{SBM}&
	\multicolumn{2}{c|}{UCI}&
	\multicolumn{2}{c}{AS}\cr\cline{2-7}
	Metrics&MAP&MRR&MAP&MRR&MAP&MRR
	\cr
	\hline
	\hline
	GCN  & 0.1894 & 0.0136 & 0.0001 &  0.0468 & 0.0019 & 0.1814 \cr
	GAT  & 0.1751 & 0.0128 & 0.0001 &  0.0468 & 0.0200 & 0.1390 \cr
	GCN-GRU  & 0.1898 & 0.0119 & 0.0114& 0.0985 & 0.0713 & 0.3388\cr
	EvolveGCN-H & 0.1947 & \underline{0.0141} & 0.0126 &  0.0899 & 0.1534 & 0.3632\cr
	EvolveGCN-O  & \textbf{0.1989} & 0.0138 & \underline{0.0270} & 0.1379 & 0.1139 & 0.2746\cr
	\hline
	DynGEM & 0.1680 & 0.0139 & 0.0209 &  0.1055 & 0.0529 & 0.1028 \cr
	dyngraph2vec \tiny{V1} & 0.0983 & 0.0079 & 0.0044 &  0.0540 & 0.0331 & 0.0698 \cr
	dyngraph2vec \tiny{V2} & 0.1593 & 0.0120 & 0.0205 &  0.0713 & 0.0711 & 0.0493  \cr
	\hline
	\hline
	\textbf{LEDG-GCN\tiny{(ours)}} &\underline{0.1960} & \textbf{0.0147} & \textbf{0.0324}   & \textbf{0.1626} & \underline{0.1932} & \textbf{0.4694}\cr
	\textbf{LEDG-GAT\tiny{(ours)}} &0.1822  & 0.0123     & 0.0261 & \underline{0.1492}  & \textbf{0.2329} & \underline{0.3835}       \cr
	\hline
	\end{tabular}}
	\vspace{-8pt}
	\label{tab:performance}
\end{table}

\subsection{Results I: Link Prediction}
The results of link prediction task are displayed in Table \ref{tab:performance}. Note that as our experiment setting in link prediction is the same as \cite{pareja2020evolvegcn}, for some of the baselines, we use the results reported in \cite{pareja2020evolvegcn}. Generally, GCN and GAT with our method significantly outperforms their vanilla versions in all datasets. Huge improvements are observed in datasets AS and UCI and our methods with GCN performs better than all the baselines significantly, which proves the effectiveness of our method in improving the performances of base models on dynamic graphs. The MAP of SBM are similar for all the supervised methods while our method with GCN is with a bit higher MRR. We observe that in this task, our method with GCN outperforms that with GAT. We argue that the reason is that GAT is more likely to overfit under such setting as generalizing to future time requires high generalization ability.

\subsection{Results II: Edge Classification}
The results of edge classification are shown in Figure \ref{edge} (left). In this task, we follow the labeling process of \cite{pareja2020evolvegcn}. A significant improvement of our method can be seen in comparison to the vanilla version of GCN and GAT. For all the three datasets, our method outperforms the baselines. The appealing results validate the effectiveness of our method towards better results in edge classification task.

\subsection{Results III: Node Classification}
\begin{figure}[t]
	\centering
	\includegraphics[width=0.47\textwidth]{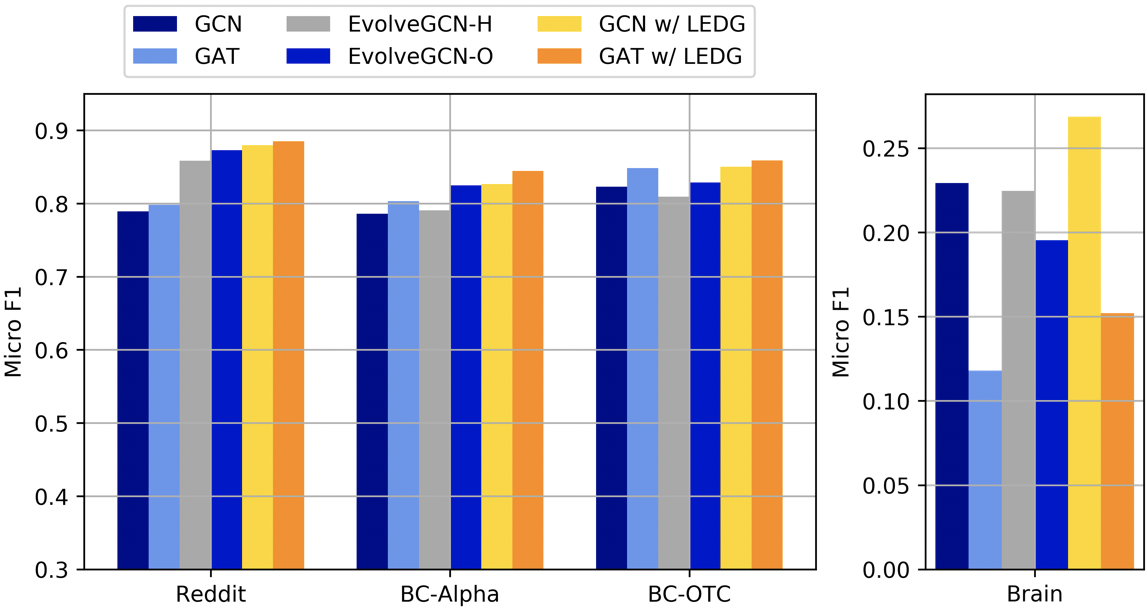}\vspace{-5pt}
	\caption{Edge classification results on datasets Reddit, BC-Alpha and BC-OTC and node classification result on Brain.}
	\vspace{-4pt}
	\label{edge}
	\vspace{-8pt}
\end{figure}

The results of node classification can be seen in Figure \ref{edge} (right). This is a 10-class classification task. The result is relatively poor as our task has no access to validation and test time. From the figure, we see that our algorithm using GCN as base model achieves the best performance. The result proves that our method can effectively work in node classification task.

\subsection{Parameter Analysis}
\begin{figure}[t]
	\centering
	\includegraphics[width=0.451\textwidth]{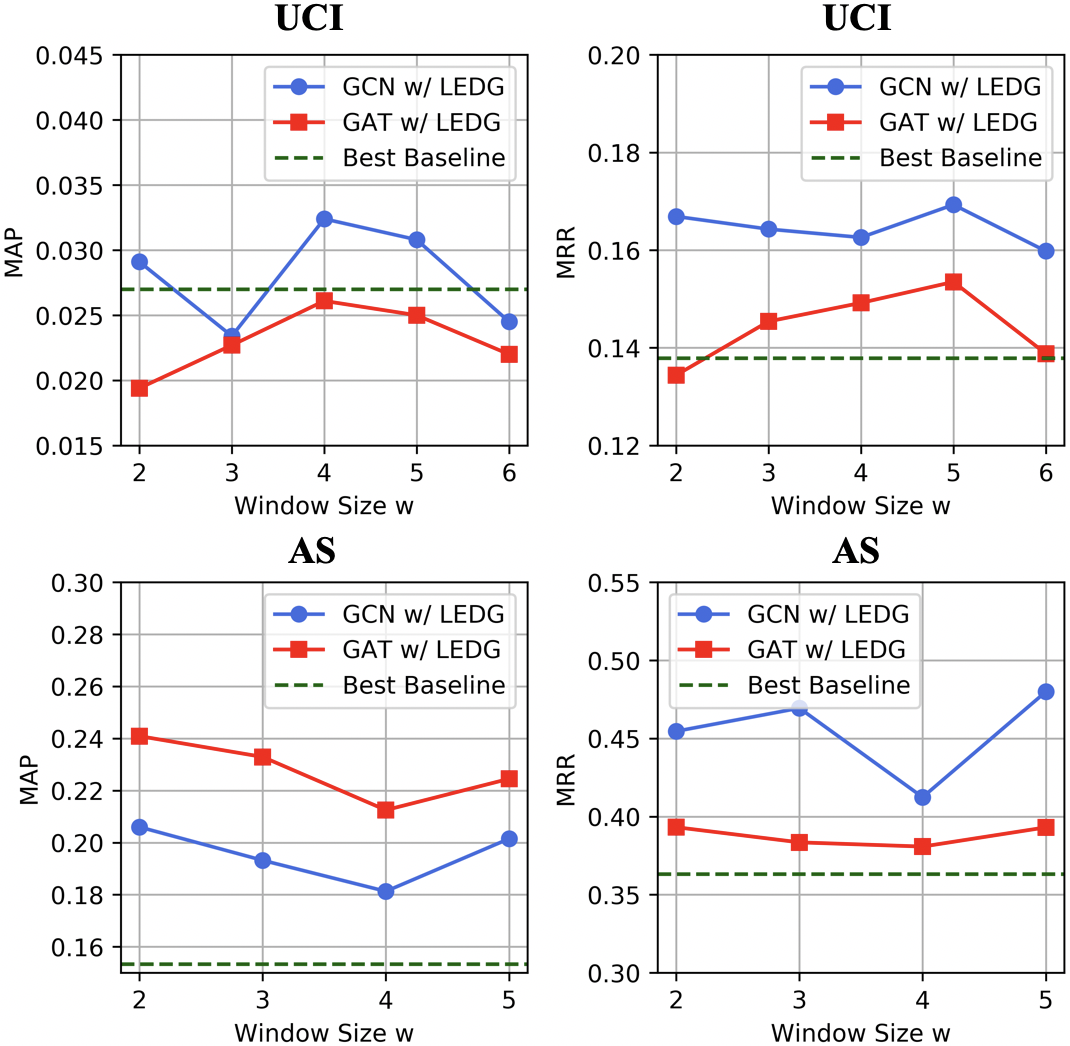} \vspace{-3.5pt}
	\caption{Link prediction performances on datasets UCI and AS with different time window size $w$.}\vspace{-4pt}
	\label{parameter}
\end{figure}

We evaluate the effect of time window size $w$ in link prediction task on datasets AS and UCI. The curves of performances are displayed in Fig. \ref{parameter} (results of size 1 are not displayed because their performances are as bad as vanilla GCN and GAT in Table \ref{tab:performance}). 
From the figure, we observe that though the time window sizes affect the performances significantly, our methods outperform the best baseline in most of the time. To highlight, our methods outperform the baseline with all the window sizes for link prediction task. 

\section{Conclusion}
We introduce a novel algorithm \textbf{LEDG} which is built on gradient-based meta-learning algorithm, for training GNNs on dynamic graphs. The algorithm learns updating strategies that have better generalization ability than RNNs. The core principle of our method is to disentangle the embeddings into time embeddings and graph intrinsic embeddings, and adapt the model parameters by time regression and down-stream tasks in a gradient-based meta-learning manner. The experiments demonstrate the effectiveness of our algorithm in training GNNs on dynamic graphs. 


\bibliographystyle{AAAI}
\bibliography{aaai22}

\begin{thebibliography}{23}
\providecommand{\natexlab}[1]{#1}

\bibitem[{{Bahdanau}, {Cho}, and {Bengio}(2015)}]{bahdanau2015neural}
{Bahdanau}, D.; {Cho}, K.; and {Bengio}, Y. 2015.
\newblock Neural Machine Translation by Jointly Learning to Align and
  Translate.
\newblock In \emph{ICLR}.

\bibitem[{Chen et~al.(2018)Chen, Xu, Wu, and Zheng}]{chen2018gc}
Chen, J.; Xu, X.; Wu, Y.; and Zheng, H. 2018.
\newblock GC-LSTM: Graph Convolution Embedded LSTM for Dynamic Link Prediction.
\newblock In \emph{CoRR}.

\bibitem[{Finn, Abbeel, and Levine(2017)}]{finn2017model}
Finn, C.; Abbeel, P.; and Levine, S. 2017.
\newblock Model-agnostic Meta-learning for Fast Adaptation of Deep Networks.
\newblock In \emph{ICML}.

\bibitem[{Finn and Levine(2018)}]{finn2017meta}
Finn, C.; and Levine, S. 2018.
\newblock Deep Representations and Gradient Descent Can Approximate Any
  Learning Algorithm.
\newblock In \emph{ICLR}.

\bibitem[{Goyal, Chhetri, and Canedo(2020)}]{goyal2020dyngraph2vec}
Goyal, P.; Chhetri, S.~R.; and Canedo, A. 2020.
\newblock Dyngraph2vec: Capturing Network Dynamics using Dynamic Graph
  Representation Learning.
\newblock \emph{Knowledge-Based Systems}, 187: 104816.

\bibitem[{Goyal et~al.(2018)Goyal, Kamra, He, and et~al.}]{goyal2018dyngem}
Goyal, P.; Kamra, N.; He, X.; and et~al. 2018.
\newblock DynGEN: Deep Embedding Method for Dynamic Graphs.
\newblock In \emph{CoRR}.

\bibitem[{Grover and Leskovec(2016)}]{grover2016node2vec}
Grover, A.; and Leskovec, J. 2016.
\newblock Node2Vec: Scalable Feature Learning for Networks.
\newblock In \emph{SIGKDD}.

\bibitem[{Hamilton, Ying, and Leskovec(2017)}]{hamilton2017inductive}
Hamilton, W.; Ying, Z.; and Leskovec, J. 2017.
\newblock Inductive Representation Learning on Large Graphs.
\newblock In \emph{NeurIPS}.

\bibitem[{{Jin} et~al.(2019){Jin}, {Zhang}, {Szekely}, and
  {Ren}}]{jin2019recurrent}
{Jin}, W.; {Zhang}, C.; {Szekely}, P.~A.; and {Ren}, X. 2019.
\newblock Recurrent Event Network for Reasoning over Temporal Knowledge Graphs.
\newblock In \emph{CoRR}.

\bibitem[{Kazemi et~al.(2020)Kazemi, Goel, Jain, Kobyzev, Sethi, Forsyth, and
  Poupart}]{kazemi2020representation}
Kazemi, S.~M.; Goel, R.; Jain, K.; Kobyzev, I.; Sethi, A.; Forsyth, P.; and
  Poupart, P. 2020.
\newblock Representation Learning for Dynamic Graphs: A Survey.
\newblock \emph{Journal of Machine Learning Research}, 21(70): 1--73.

\bibitem[{{Kipf} and {Welling}(2017)}]{kipf2016semi}
{Kipf}, T.~N.; and {Welling}, M. 2017.
\newblock Semi-Supervised Classification with Graph Convolutional Networks.
\newblock In \emph{ICLR}.

\bibitem[{Li et~al.(2019)Li, Han, Cheng, Su, Wang, Zhang, and
  Pan}]{li2019predicting}
Li, J.; Han, Z.; Cheng, H.; Su, J.; Wang, P.; Zhang, J.; and Pan, L. 2019.
\newblock Predicting Path Failure In Time-Evolving Graphs.
\newblock In \emph{SIGKDD}.

\bibitem[{{Manessi}, {Rozza}, and {Manzo}(2020)}]{manessi2020dynamic}
{Manessi}, F.; {Rozza}, A.; and {Manzo}, M. 2020.
\newblock Dynamic Graph Convolutional Networks.
\newblock \emph{Pattern Recognition}, 97: 107000.

\bibitem[{Mikolov et~al.(2013)Mikolov, Sutskever, Chen, Corrado, and
  Dean}]{mikolov2013distributed}
Mikolov, T.; Sutskever, I.; Chen, K.; Corrado, G.~S.; and Dean, J. 2013.
\newblock Distributed Representations of Words and Phrases and their
  Compositionality.
\newblock In \emph{NeurIPS}.

\bibitem[{Pareja et~al.(2020)Pareja, Domeniconi, Chen, and
  et~al.}]{pareja2020evolvegcn}
Pareja, A.; Domeniconi, G.; Chen, J.; and et~al. 2020.
\newblock EvolveGCN: Evolving Graph Convolutional Networks for Dynamic Graphs.
\newblock In \emph{AAAI}.

\bibitem[{Pei et~al.(2019)Pei, Wei, Chang, Lei, and Yang}]{Geom-GCN}
Pei, H.; Wei, B.; Chang, K. C.-C.; Lei, Y.; and Yang, B. 2019.
\newblock Geom-GCN: Geometric Graph Convolutional Networks.
\newblock In \emph{ICLR}.

\bibitem[{Perozzi, Al-Rfou, and Skiena(2014)}]{perozzi2014deepwalk}
Perozzi, B.; Al-Rfou, R.; and Skiena, S. 2014.
\newblock DeepWalk: Online Learning of Social Representations.
\newblock In \emph{SIGKDD}.

\bibitem[{Rossi and Ahmed(2015)}]{nr}
Rossi, R.~A.; and Ahmed, N.~K. 2015.
\newblock The Network Data Repository with Interactive Graph Analytics and
  Visualization.
\newblock In \emph{AAAI}.

\bibitem[{Sankar et~al.(2020)Sankar, Wu, Gou, Zhang, and
  Yang}]{sankar2020dysat}
Sankar, A.; Wu, Y.; Gou, L.; Zhang, W.; and Yang, H. 2020.
\newblock DySAT: Deep Neural Representation Learning on Dynamic Graphs via
  Self-Attention Networks.
\newblock In \emph{WSDM}.

\bibitem[{Skarding, Gabrys, and Musial(2020)}]{skarding2020foundations}
Skarding, J.; Gabrys, B.; and Musial, K. 2020.
\newblock Foundations and Modelling of Dynamic Networks using Dynamic Graph
  Neural Networks: A Survey.
\newblock In \emph{CoRR}.

\bibitem[{{Velickovic} et~al.(2018){Velickovic}, {Cucurull}, {Casanova}, and
  et~al.}]{velivckovic2017graph}
{Velickovic}, P.; {Cucurull}, G.; {Casanova}, A.; and et~al. 2018.
\newblock Graph Attention Networks.
\newblock In \emph{ICLR}.

\bibitem[{{Xu} et~al.(2019){Xu}, {Cheng}, {Luo}, and et~al.}]{xu2019adaptive}
{Xu}, D.; {Cheng}, W.; {Luo}, D.; and et~al. 2019.
\newblock Adaptive Neural Network for Node Classification in Dynamic Networks.
\newblock In \emph{ICDM}.

\bibitem[{Xu et~al.(2018)Xu, Hu, Leskovec, and Jegelka}]{GIN}
Xu, K.; Hu, W.; Leskovec, J.; and Jegelka, S. 2018.
\newblock How Powerful are Graph Neural Networks?
\newblock In \emph{ICLR}.

\end{thebibliography}

\end{document}